\documentclass[a4paper,fleqn]{cas-dc}

\usepackage[numbers]{natbib}
\usepackage{epstopdf}
\usepackage{color}
\usepackage{booktabs} 
\usepackage{hyperref}
\usepackage{verbatim}
\def\tsc#1{\csdef{#1}{\textsc{\lowercase{#1}}\xspace}}
\tsc{WGM}
\tsc{QE}
\tsc{EP}
\tsc{PMS}
\tsc{BEC}
\tsc{DE}

\begin{document}
\let\WriteBookmarks\relax
\def\floatpagepagefraction{1}
\def\textpagefraction{.001}
\shorttitle{}
\shortauthors{Z. Feng et~al.} 

\title [mode = title]{VS-CAM: Vertex Semantic Class Activation Mapping to Interpret Vision Graph Neural Network}                      



\author[1]{\textcolor[RGB]{0,0,1}{Zhenpeng Feng}}
\cormark[1]
\ead{zpfeng_1@stu.xidian.edu.cn}
\address[1]{School of Electronic Engineering, Xidian University, Xi'an, China}  

\author[1]{\textcolor[RGB]{0,0,1}{Xiyang Cui}}

\author[1]{\textcolor[RGB]{0,0,1}{Hongbing Ji}}

\author[1]{\textcolor[RGB]{0,0,1}{Mingzhe Zhu}}

\author[2]{\textcolor[RGB]{0,0,1}{Ljubi\v{s}a Stankovi\'c}}
\address[2]{Faculty of Electrical Engineering, University of Montenegro, Podgorica, Montenegro}

\cortext[cor1]{Corresponding author: Zhenpeng Feng} 



\begin{abstract}
Graph convolutional neural network (GCN) has drawn increasing attention and attained good performance in various computer vision tasks, however, there lacks a clear interpretation of GCN's inner mechanism. For standard convolutional neural networks (CNNs), class activation mapping (CAM) methods are commonly used to visualize the connection between CNN's decision and image region by generating a heatmap. Nonetheless, such heatmap usually exhibits semantic-chaos when these CAMs are applied to GCN directly. In this paper, we proposed a novel visualization method particularly applicable to GCN, Vertex Semantic Class Activation Mapping (VS-CAM). VS-CAM includes two independent pipelines to produce a set of semantic-probe maps and a semantic-base map, respectively. Semantic-probe maps are used to detect the semantic information from semantic-base map to aggregate a semantic-aware heatmap. Qualitative results show that VS-CAM can obtain heatmaps where the highlighted regions match the objects much more precisely than CNN-based CAM. The quantitative evaluation further demonstrates the superiority of VS-CAM.
\end{abstract}



\begin{keywords}
keywords-1 graph neural network\sep
keywords-2 interpreting neural network\sep 
keywords-3 class activation mapping\sep 
\end{keywords}

\maketitle  

\section{Introduction}\label{sec:1}
Convolutional neural network (CNN) used to be the most powerful structure in various computer vision tasks, like image classification~\cite{AlexNet, ResNet}, object detection~\cite{Yolo, Zhuzhigang, TS4Net}, semantic segmentation \cite{UNet, TPAMI, TIP}, etc. Recently, numerous vision transformer (ViT) architectures based on self-attention are introduced and rapidly take dominance in vision tasks \cite{ViT, ViT2, UTRAD}. CNNs utilize sliding
kernels over many grids of pixels in the Euclidean space, thus CNN retains two important inductive biases for images: locality and shift-invariance. The recent ViT models treat the image as a sequence of patches to establish long-range dependency between two patches but forsake locality and shift-invariance. However, the objects are
usually not in a regular quadratic shape, thus the commonly-used grid or sequence structures in CNN and ViT are inflexible to process them. 

Different from regular grid or sequence representation, Graph neural network (GNN) processes the image in a more flexible way \cite{ViG}.  
An object can be deemed as a constitution of multiple components, e.g., a bird can be roughly divided into head, body, and wings. These parts are viewed as vertices in graph signal processing and edges are used to represent the connections among vertices.
By aggregating and updating the information of a vertex with its neighboring vertices in the graph, GNN is able to strengthen the connections among vertices relevant to the object while suppressing those object-irrelevant.  
Furthermore, graph is a generalized data structure that grid and sequence can be viewed as a special case of graph \cite{StankovicGCN, StankovicGSP}. 
In computer vision, GNN is mainly applied in image classification, scene graph generation, and action recognition. T. Kipl et al. proposed an efficient variant of convolutional neural networks which
operate directly on graph-structured
data and outperforms related methods significantly in semi-supervised classification tasks \cite{GCN}. Scene graph generation aims to parse the input image into a graph with the objects
and their relation by combining the object detector and GCN \cite{SceneGrapj}. By
processing the naturally formed graph of linked human joints, GCN is utilized in human action
recognition task \cite{actionGCN, AST}. However, GCN can only process specific visual tasks with naturally constructed graphs. For general applications in computer vision, K. Han et al. recently proposed vision GNN (ViG) backbone that directly
processes the image data \cite{ViG}.  
ViG attains competitive performance in accuracy and computation cost compared with CNN and ViT in image classification and object detection tasks. 

\begin{figure}[t]
	\centering
	{\includegraphics{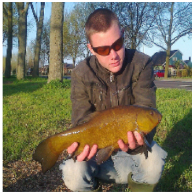}}
	{\includegraphics{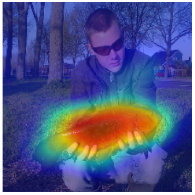}}
	{\includegraphics{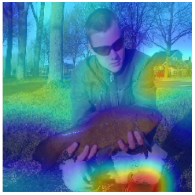}}
	{\includegraphics{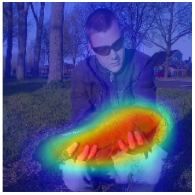}}
	
	{\includegraphics{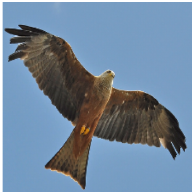}}
	{\includegraphics{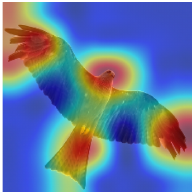}}
	{\includegraphics{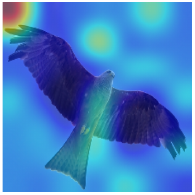}}
	{\includegraphics{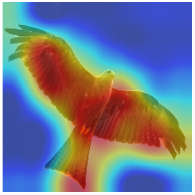}}
	
	{\includegraphics{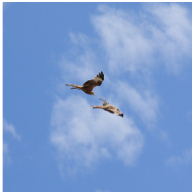}}
	{\includegraphics{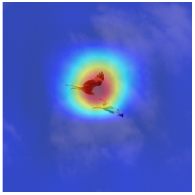}}
	{\includegraphics{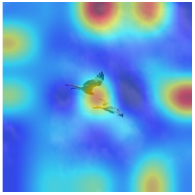}}
	{\includegraphics{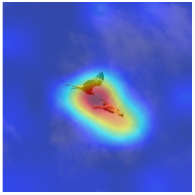}}
	
	{\includegraphics{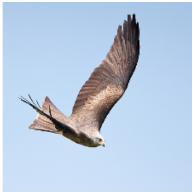}}
	{\includegraphics{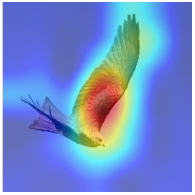}}
	{\includegraphics{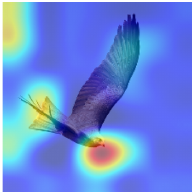}}
	{\includegraphics{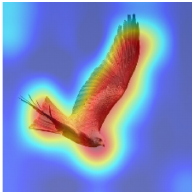}}
	\caption{Visualization of heatmaps on ILSVRC dataset. Original image (first-column). CNN-CAM (second-column). Basic CAM for ViG (third-column). ViG-CAM (fourth-column).}\label{fig: camcomparison}
\end{figure}

Despite the success of ViG in computer vision, there lacks enough interpretation of what ViG has learned inside the features in each layer to make a correct decision. 
Therefore, it is highly desirable and necessary to understand and interpret what exactly ViG learned, especially for applications where interpretability is essential (e.g., medical diagnosis and autonomous driving) \cite{Xgradcam}. 
B. Zhou proposed class activation mapping (CAM) to utilize the activation maps from the
last convolution layer to generate semantic-aware saliency heatmap to visualize CNN's mechanism \cite{CAM}.  
However, CAM suffers from a severe underestimation of object regions
because the discriminative regions activated through the classification
models are often much smaller than the objects’ actual
extent. Worse still, the heatmap is almost semantic-chaos with ViG when CAM is used directly, as shown in Fig.~\ref{fig: camcomparison}.

In this paper, we propose a vertex-semantic class activation mapping (VS-CAM), making the first attempt for interpreting ViG's inner mechanism. VS-CAM includes two independent pipelines: one generates a semantic-base map only by combining the gradient of classification score to the output of a layer in ViG; the other performs a set of semantic-probe maps only by features in a layer to detect the class-discriminative regions in the semantic-base map. Finally, the semantic-base map is coupled with each semantic-probe map to form the current element in VS-CAM heatmap. 

The contributions of this paper are as follows:
\begin{itemize}
	\item We propose the vertex-semantic class activation mapping
	(VS-CAM), as the first attempt for interpreting the mechanism of
	visual graph neural network in image classification tasks. VS-CAM requires no backpropagation where two pipelines are implemented independently by aggregating the gradient and measuring the similarity of vertices, respectively
	\item VS-CAM outperforms other previous
	methods on a commonly-used image benchmark by fully exploiting both local-range features and long-range vertices dependencies in the visual graph neural network.
\end{itemize}

The rest of this paper is organized as follows. Section~\ref{sec:2} introduces the basic knowledge of ViG and CAM. Section~\ref{sec:3}  describes how to generate saliency heatmaps by VS-CAM in detail. In Section~\ref{sec:4},
various experiments are implemented to demonstrate the validity of VS-CAM and further interpret ViG from several aspects. Section~\ref{sec:5} concludes this paper.


\section{Related Work}\label{sec:2}

\subsection{Vision Graph Neural Network}
Vision graph neural network is comprised of three parts: stem, backbone, and classifier \cite{ViG}. Stem includes three normal convolutional layers followed by nonlinear activation. Backbone is a stack of ViG blocks (i.e., the basic unit in ViG) as well as three normal convolutional layers for downsampling. The classifier is a multi-layer perceptron (MLP) which outputs the classification score of each class. Next, we will start from constructing the graph topology of an image to form a ViG block.

\begin{figure}[t]
	\centering
	{\includegraphics[]{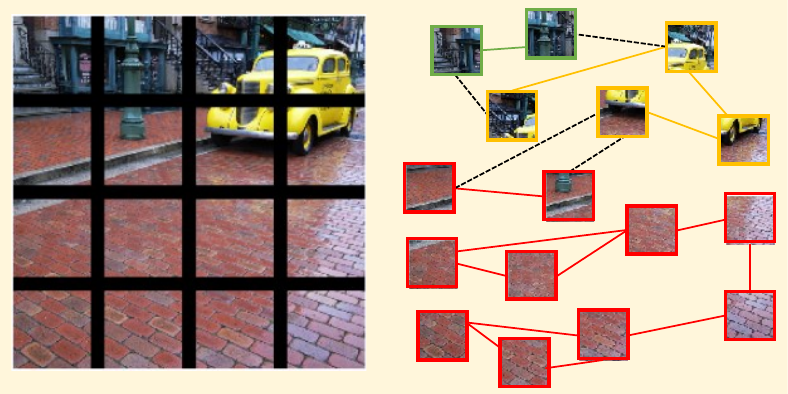}}
	\caption{ The graph topology of an image. An image is firstly divided into $N$ patches ($N=16$ is only for a clear exhibition) (left). The topology of the graph constructed in this image (right). Each path is viewed as a vertex in the graph and its two nearest vertices are chosen as neighbors. Three pathes include three objects: car, building, and street, which are marked by yellow, green, and red squares, respectively. The dotted line denotes two vertices with different objects are connected.}\label{fig:graph}
\end{figure}

\noindent \textbf{Graph Topology of Image} Graph signal processing views the input data as a set of vertices, $\mathbf{V}$, connected by the set of edges, $\mathbb{E}$, which is termed topology of a graph. In our analysis, an input image $\mathbf{I}\in \mathbb{R}^{M\times M\times D}$ of size $M \times M$ pixels with $D$ channels, is firstly divided uniformly into $N$ patches, $\mathbf{x}_1, \mathbf{x}_2, \dots,\mathbf{x}_N$, whose size is $(M/\sqrt{N}\times M/\sqrt{N})$, as illustrated in Fig. \ref{fig:graph} (left). All patches form a set of features, written in matrix form as 
$$\mathbf{X}=\{\mathbf{x}_1, \mathbf{x}_2, \dots,\mathbf{x}_N\},$$ 
with $\mathbf{x}_i\in \mathbb{R}^{M/\sqrt{N}\times M/\sqrt{N}\times D}$.
Note for a large input image, the patches can be down-sampled to save computation cost. When a large number of patches (a small size of each patch) is used, then a single pixel, which is obtained by averaging over all pixels within one patch, can be used.  In this simplified case, the set of patches, $\mathbb{X}$, consists of $N$ vectors $\mathbf{x}_i$, whose dimension is $D$.

For graph-based analysis, the patches $\mathbf{x}_i$, $i=1,2,\dots,N$, are associated with vertices, labeled by $v_i$, $i=1,2,\dots,N$, forming the set $\mathbb{V}=\{v_1,v_2,\dots,v_N\}$.  The vertices should be connected in order to form the domain for data (i.e., patches in our case) processing. In classical data processing, we would connect the vertices according to their spatial order. In graph-based data processing, it is more common to connect the vertices taking into account the similarity of associated patches (data). Since the patches are small sub-images, we can use one of the numerous methods developed for measuring image similarity (for an overview see \cite{StankovicGCN}). The simplest method is based on the Euclidean distance of the patches intensity defined by 
\begin{align}
	r_{ij}=\mathrm{distance}(\mathbf{x}_{i}, \mathbf{x}_{j}) = \|\mathbf{x}_{i} - \mathbf{x}_{j}\|_{2},
\end{align}
where $j = 1, 2, \dots, N, j\neq i$.
In the simplified case, when each patch is one pixel, $\|\cdot\|_{2}$ denotes the standard norm-two, while in the case when the patches are small images, this is the Frobenius norm form. 

The distances $r_{ij}$ are assigned to the edge weights. In order to reduce the number of edge weights and avoid graph over-connectivity, various mappings of the calculated distance or thresholds are used. Here we use a very simplified form where the edges for only $K$ the strongest connected vertices are used in their normalized form. It means that the edge weights are assigned to the graph, $G(\mathbb{V},\mathbb{E})$, as
\begin{align}
	A(i,j) =\begin{cases} 1, \ \ \text{if} \  \ (v_i,v_j) \in \mathbb{E}, \\
		0,  \ \ \text{elsewhere},
	\end{cases}
\end{align}
where the set $\mathbb{E}$ of pairs of vertices $(v_i,v_j)$ that contains $K$ pairs whose distance is minimum, that is  
\begin{align}
	\mathbb{E}=\{(v_i,v_j) |  \ \ \  \arg \min_{(v_i,v_j), \ K,\ j \ne i}{	\mathrm{distance}(\mathbf{x}_{i}, \mathbf{x}_{j}) } \},
\end{align}  
where $\min \limits_{ K}(\cdot)$ denotes finding the first $K$ minimum values.
After that, we can add an edge $A_{ji} = 1$ directed from $v_j$ to $v_i$ for all $v_j\in\mathbb{K}$, otherwise, $A_{ji} = 0$. This kind of weight matrix (with elements being either 1 or 0) is called adjacency matrix.
Now the graph
$G(\mathbb{V},\mathbb{E})$ is obtained. For the considered image, the resulting graph is presented in Fig. \ref{fig:graph} (right).

\noindent \textbf{Graph Convolution}
The data is processed as a graph using its values at the considered vertex and its neighboring vertices. The data at a vertex $v_i$ is denoted by $\mathbf{x}_i$. For each vertex, we consider $D$ channels, meaning that the input data at vertex $v_i$ and at channel $d$, $d=1,2,\dots,D$, are denoted by $x_i(d)$. Data in one specific channel, over all vertices, can be denoted by a vector 
\begin{align}
	\mathbf{x}(d)=[x_1(d), \ x_2(d),  \ \dots, x_N(d)]^T.
\end{align}
The output of a first-order system of a signal (one channel) on graph is 
\begin{align}
	\mathbf{o}(d)=w_0(d)\mathbf{x}(d)+w_1(d)\mathbf{A}\mathbf{x}(d), d=1,2,\dots,D,
\end{align}  
where $w_0(d)$ and $w_1(d)$ are the system parameters. The output has two parts, one the signal itself, $\mathbf{x}(d)$, and the other part, $\mathbf{A}\mathbf{x}(d)$, called  aggregate of data within the established similar vertices, 
\begin{align}
	\mathbf{y}(d)=\mathbf{A}\mathbf{x}(d), \text{ for } d=1,2,\dots,D.
\end{align}  
When averaging is the aim, then the aggregated value should be divided by the number of non-zero values in one row of the matrix $\mathbf{A}$. This is done by using the diagonal degree matrix, $\mathbf{D}$, 
\begin{align}
	\mathbf{y}(d)=\mathbf{D}^{1/2}\mathbf{A}\mathbf{D}^{1/2}\mathbf{x}(d), \text{ for } d=1,2,\dots,D.
\end{align} 
The aggregated data can again be reorganized for each vertex as 
\begin{align}
	\mathbf{y}_i=[y_i(1), y_i(2),\dots,y_i(D)]^T, \text{ for } n=1,2,\dots,N.
\end{align} 
The signal and its aggregation can be arranged as 
\begin{align}
	\mathbf{g}_i=[\mathbf{x}_i, \ \ \mathbf{y}_i] \text{ for } i=1,2,\dots,N.
\end{align}
The output of this graph convolution layer is then obtained by a linear combination of both, the signal and its aggregate.

The averaging (mean aggregate) is just one possible form of combining the output from the graph CNN \cite{GCN}. Other forms are, for example, a max-pooling aggregator \cite{hamilton2017inductive, qi2017pointnet, wang2019dynamic} and  attention aggregator \cite{velivckovic2017graph}. Instead of the mean aggregate, we will use max-pooling of the feature difference aggregator \cite{maxconv}, without learnable parameters, to emphasize the maximum difference in the data at the considered vertex and its neighboring vertices defined by the matrix $\mathbf{W}$. This kind of aggregate is as
\begin{align}
	\mathbf{g}_i=[\mathbf{x}_i, \ \ \max_j\{\mathbf{x}_i-\mathbf{x}_j\},
	\text{ for } (v_i,v_j)\in \mathbb{K} ],
\end{align}
where $i=1,2,\dots,N$. Notice that in this kind of aggregation no learnable parameters are used. The original signal and aggregated signal should be combined in the graph convolution (updated), using learnable parameters in matrix $\mathbf{W}$, as
\begin{align}
	\mathbf{h}_i=\mathbf{g}_i\mathbf{W}.
\end{align}  
For a channel, the update is $\mathbf{W}=[w(0), \ w(1)]^T$. If we use $D$ channels, then we have one $\mathbf{W}_c$, meaning that the total number of parameters is $C$. The data at one vertex can be split over channels in several groups, $\mathbf{g}_i=[\mathbf{g}^1_i, \  \mathbf{g}^2_i, \ \dots, \mathbf{g}^H_i]$, to add more flexibility in these learnable parameters, 
\begin{align}
	\mathbf{h}_i=[\mathbf{g}^1_i\mathbf{W}_1, \  \mathbf{g}^2_i\mathbf{W}_2, \ \dots, \ \mathbf{g}^H_i\mathbf{W}_H].
\end{align}

The above graph convolution processing can be denoted as 
\begin{align}\mathbf{H}= \mathrm{GraphConv}(\mathbf{X})\end{align},
including both aggregation and update operation, and  $\mathbf{H}=[\mathbf{h}_1, \mathbf{h}_2, \dots,\mathbf{h}_N]$.

\noindent \textbf{ViG Block}
To alleviate the over-smoothing phenomenon commonly appearing in previous GCNs, ViG introduces a module called Grapher including fully connected layers before and after the graph convolution
to project the vertex features into the same domain
and increase the feature diversity \cite{ViG}. A common nonlinear activation
function is inserted after graph convolution:
\begin{align}
	\mathbf{Y} = \mathrm{Grapher}(\mathbf{V})= \sigma(\mathrm{GraphConv}(\mathbf{V}\mathbf{W}_{in}))\mathbf{W}_{out}
\end{align}
where $\mathbf{Y}\in\mathbb{R}^{N\times D}$ is the output of $Grapher$ module.
To further encourage the feature transformation capacity and relief the over-smoothing phenomenon,
ViG utilizes a feed-forward network ($\mathrm{FFN}$) on each vertex. The $\mathrm{FFN}$ module is a simple multi-layer
perceptron with two fully-connected layers \cite{ViG}:
\begin{align}
	\mathbf{Z} = \mathrm{FFN}(\mathbf{Y})= \sigma(\mathbf{Y}\mathbf{W}_1)\mathbf{W}_2 + \mathbf{Y},
\end{align}
where $\mathbf{Z}\in\mathbb{R}^{N\times D}$, $\mathbf{W}_1$ and $\mathbf{W}_2$ are the weights of fully-connected layers. A stack of $Grapher$ modules and $\mathrm{FFN}$ modules constitute the  ViG block which is the basic unit in ViG backbone \cite{ViG}. 

Then the entire ViG architecture, $	\Im$, can be formed by concatenating stem, backbone, and classifier. For classification task, the probability score of each class, $\mathbf{p} \in \mathbb{R}^{C}$ where $C$ denotes the number of classes of dataset, is the output of the last layer in the classifier as
\begin{equation}
	\mathbf{p}(i) = \Im(\mathbf{I}),  \text{ for } i = 1, 2, \dots, C.
\end{equation}
The detailed parameters of ViG used in this paper can be found in the Section~\ref{sec:4}.

\subsection{Class Activation Mapping}
An important issue in interpreting CNN is to explain why classification CNN learned from the input data to make a correct prediction \cite{ProbeFeature, CleverHans}. To show what CNN looks for in the input image, numerous CAM methods are proposed to visualize CNN’s decision using feature maps in deep layers. B. Zhou et al. proposed the original CAM method which forms a saliency heatmap by linearly combining feature maps at the last layer \cite{CAM}.
The weight of each feature map is determined by the last layer’s fully-connected weights corresponding to an object class. In this case, CAM is only applicable to those CNNs with global-average pooling as the last layer.
To avoid modifying CNN's structure, Selvaraju et al.
then proposed Grad-CAM to visualize an arbitrary CNN for classification by weighting the
feature maps in $l$-th convolutional layer using gradients \cite{GradCAM}. 
However, the highlighted regions generated by Grad-CAM are usually much smaller than the object. To provide a complete highlighted region, some modified CAMs are proposed, like Grad-CAM++ \cite{GradCAMplus}, Ablation CAM \cite{AblationCAM}, Score CAM \cite{ScoreCAM}, Self-Matching CAM \cite{SelfMatchingCAM, SCSMCAM}, etc. In this work, we mainly focus on Grad-CAM for its generality and simplicity.
The Grad-CAM heatmap, $\mathbf{M}^{c}_{grad}$, is formulated as:
\begin{align}
	&\mathbf{M}_{grad}^{c}  =  \sum_{d}\alpha_{d}^{c} \mathbf{F}_{d}^{l}, \notag \\
	&\alpha^{c}_{d}  =  \sum_{i}\sum_{j} \frac{\partial \mathbf{p}_{c}}{\partial \mathbf{F}^{l}_{d}(i,j)},
\end{align}
where $\mathbf{p}_c$ is the classification score of $c$ class for a given input image, the weight, $\alpha^{c}_{d}$, is the element-wise summation of the partial gradient of $\mathbf{p}_c$ to the $d$-th feature map in $l$ -th layer, $\mathbf{F}^{l}_{d}$.  Note that this Grad-CAM is effective in various CNN models, while the heatmap usually shows agnostic semantics when Grad-CAM is applied to ViG directly.

\section{Methodology}\label{sec:3}
\begin{figure*}[t]
	\centering
	{\includegraphics[width=18.5cm]{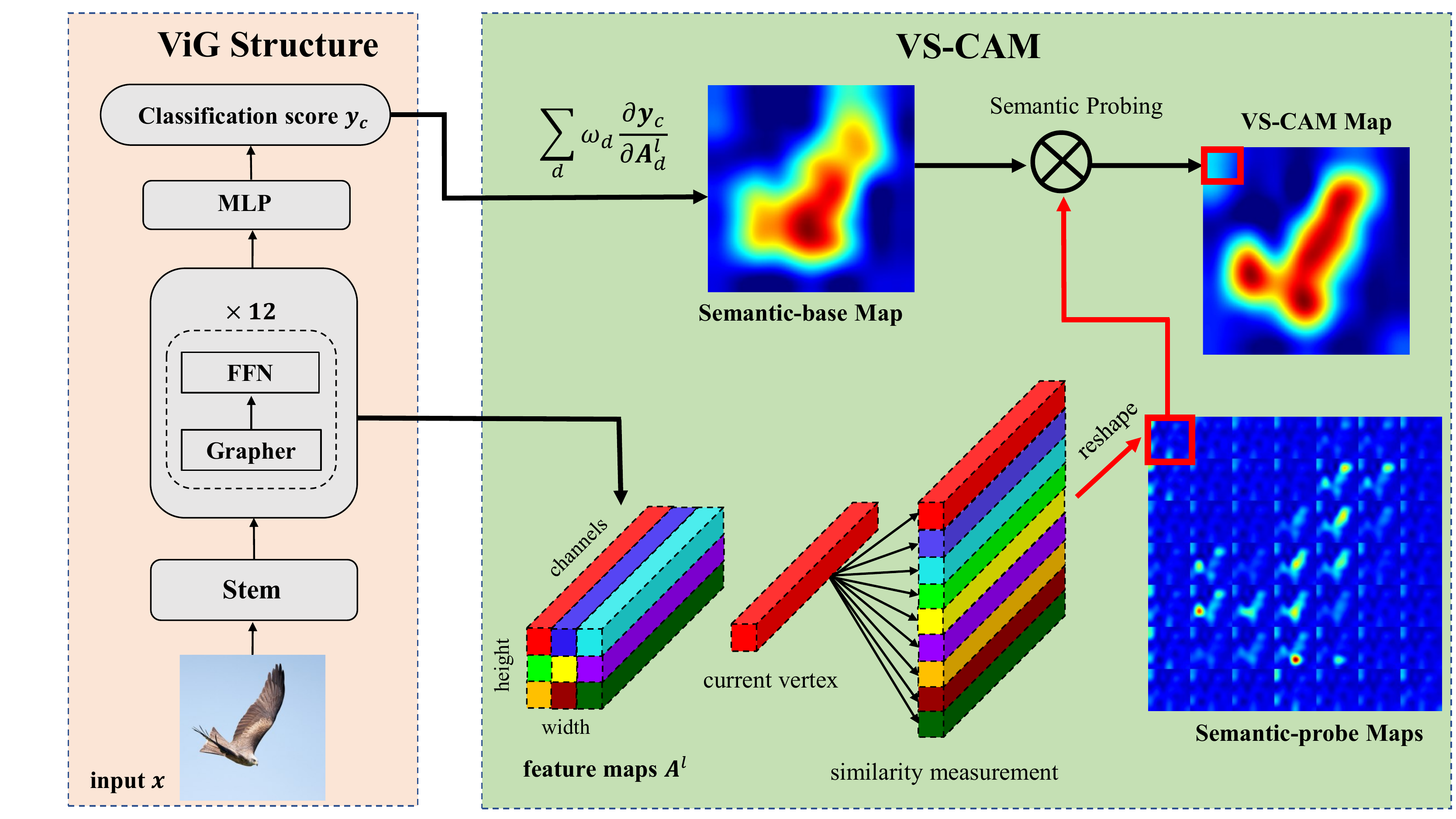}}
	\caption{VS-CAM framework, which consists of a ViG model, a branch of Semantic-base Map, a branch of Semantic-probe Maps, and a semantic probing operator. }\label{fig:flowchart}
\end{figure*}
To avoid semantic-chaos heatmaps by using common Grad-CAM to ViG in Fig.~\ref{fig: camcomparison},
we propose the VS-CAM method to generate saliency heatmaps upon the pre-trained ViG in Fig.~\ref{fig:flowchart}.
To fulfill semantic-aware region in heatmaps, we introduce a semantic-base map  and a set of semantic-probe maps to detect hidden class-discriminative regions in semantic-base map. 
Each semantic-probe map is element-wise multiplied with the
semantic-base map to obtain the value of the current element in the generated VS-CAM heatmap.

\noindent\textbf{Semantic-probe maps}:
Suppose $\mathbf{F}^{l}$ $\in$ $ \mathbb{R}^{W^{l}\times H^{l}\times D^{l}}$, is the feature maps of the Grapher in the $l$-th block of ViG. Here we firstly introduce a set of the similarity matrices, $\mathbf{S}_{w,h}$, formulated as follows: 
\begin{align}
	\mathbf{S}_{1,1} (i,j) &= (\mathbf{F}^{l}(1,1), ~\mathbf{F}^{l}(i,j) ) \notag \\ 
	\mathbf{S}_{1,2} (i,j) &= (\mathbf{F}^{l}(1,2) , ~  \mathbf{F}^{l}(i,j) ) \notag \\
	&\cdots	\notag \\
	\mathbf{S}_{W^{l},H^{l}} (i,j) &= (\mathbf{F}^{l}(W^{l},H^{l}) , ~  \mathbf{F}^{l}(i,j))
	\label{eq:similarity_matrix}.
\end{align}
where $(\mathbf{a}, \mathbf{b})$ denotes inner production operator of vector $\mathbf{a}$ and $\mathbf{b}$, $\mathbf{F}^{l}(i,j)$ is the $i$-th and $j$-th vector in length of $D$, $i=1,2, \dots, W^{l}$, and  $j=1,2, \dots, H^{l}$.

\noindent\textbf{Semantic-base maps}:
Suppose $c$ is the ground truth label of a given image, we only preserve the prediction score in $c$-th element in $\mathbf{p}$ while the rest elements are set to zero to obtain a $\mathbf{p}_c\in \mathbb{R}^{C}$ as
\begin{align}
	\mathbf{p}_c(i) =\begin{cases} \mathbf{p}(i), \ \ \text{if} \  \ i=c, \\
		0,  \ \ \text{elsewhere},
	\end{cases}
\end{align}
Here we only use $\frac{\partial{\mathbf{p}_{c}}}{\partial{\mathbf{A}^{l}}}$,  the gradient of $\mathbf{p}_c$ with respect to  $\mathbf{F}^{l}$, to form semantic-base maps, $\mathbf{Q}^{c}$, as follows:
\begin{align}
	\mathbf{Q}^{c} &=  \sum_{d}{\omega_{d} \frac{\partial{\mathbf{p}_{c}}}{\partial{\mathbf{F}^{l}_{d}}}}, ~d=1, 2, \dots, D, \notag \\
	\omega_{d} &= \sum_{i}{\sum_{j}{\frac{\partial{\mathbf{p}_{c}}}{\partial{\mathbf{F}^{l}_{d}}}}},~ i=1, 2, \dots, W^{l},~ j=1, 2, \dots, H^{l}.
\end{align}	

\noindent\textbf{VS-CAM}:
VS-CAM, $\mathbf{M}^{c}_{vs}$, is obtained by coupling semantic-base maps and semantic-probe maps as:
\begin{align}
	\mathbf{M}^{c}_{vs}(i,j) = \Gamma^{W\times H}( \sum_{m}\sum_{n}{{\mathbf{S}_{i,j}(m,n)\odot \mathbf{Q}^{c}}}),
\end{align}	
where $\odot$ denotes Hadamard product operator, $\Gamma^{W\times H}$ denotes resizing  $\mathbf{S}_{i,j}\odot \mathbf{F}$ ($\mathbb{R}^{W^{l} \times H^{l}}$) to the shape of input image ($\mathbb{R}^{W \times H}$).

\section{Experiments}\label{sec:4}

\subsection{Experimental Setup}
\noindent\textbf{Dataset:} The ViG model is
trained and validated on a commonly used benchmark, i.e., ILSVRC. In ILSVRC, there
are around 1.2 million images 1000 categories for
training, and 50 thousand images 1000 categories for validation. All the saliency heatmaps are generated from pre-trained CNN and ViG models with the above two datasets. 

\noindent\textbf{Network Structure:}
\begin{table}[tbp]
	\centering
	\caption{The architecture of ViG. \label{tab:vig}}
	\begin{tabular}{ccccc}
		\toprule  
		Stage&Output shape&Layer&\\ 
		\midrule  
		Stem &$W/4\times W/4$	& Conv2d$\times3$ & \\
		\hline
		&$W/4\times W/4$	&   Block($D=48$)$\times2$  & \\
		&$W/8\times W/8$ &  Conv2d &  \\
		&$W/8\times W/8$ 	&Block($D=96$)$\times2$  & \\	
		Backbone&$W/16\times W/16$  	& Conv2d & \\	
		&$W/16\times W/16$ 	&Block($D=240$)$\times6$   & \\	
		&$W/32\times W/32$ 	& Conv2d  &  \\	
		&$W/32\times W/32$	& Block($D=384$)$\times2$ & \\	
		\hline
		Classifier&$1\times1$ 	& Pooling \& MLP &  \\	
		\bottomrule  
	\end{tabular}
\end{table}	
\cite{ViG} built four versions of ViG architecture with different model sizes, i.e., ViG-Ti, S, M, and B. In this paper, we only focus on the smallest model, ViG-Ti, with $10.7$M parameters. The detailed architecture of ViG-Ti is shown in Table~\ref{tab:vig}. ViG-Ti can reach $78.5\%$ classification accuracy while MobileNet-V3-large ($7.5$M parameters), ResNet-18 ($12$M parameters) \cite{MobileNetV3, ResNet, ResNet2} and PVT-Tiny ($13.2$M parameters) \cite{PViT} only reach $	
74.0$ , $70.6\%$ and $75.1\%$, respectively \cite{ViG}.
As a comparison, we adopt a light and good-performance CNN model, MobileNet-V3-large, in this paper. More details of MobileNet-V3-large can be found in \cite{MobileNetV3}.

\subsection{Performance of Discriminative Localization}
\begin{figure*}[t]
	\centering
	{\includegraphics[]{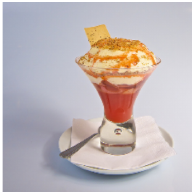}}
	{\includegraphics[]{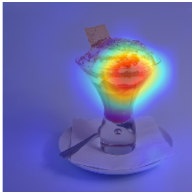}}
	{\includegraphics[]{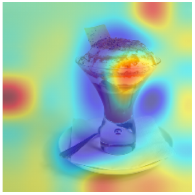}}
	{\includegraphics[]{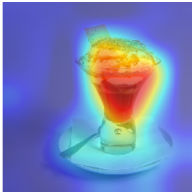}}
	{\includegraphics[]{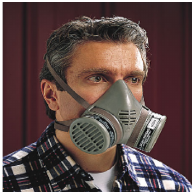}}
	{\includegraphics[]{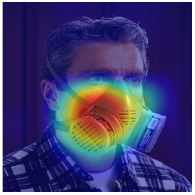}}
	{\includegraphics[]{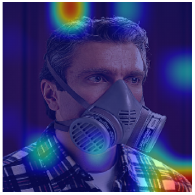}}
	{\includegraphics[]{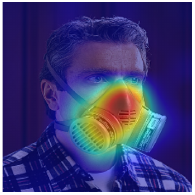}}
	
	{\includegraphics[]{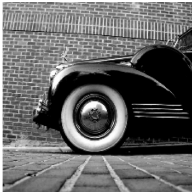}}
	{\includegraphics[]{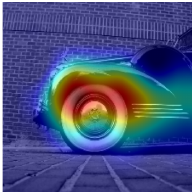}}
	{\includegraphics[]{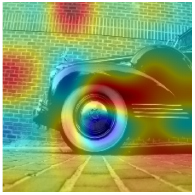}}
	{\includegraphics[]{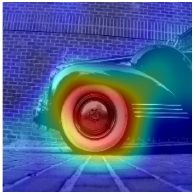}}
	{\includegraphics[]{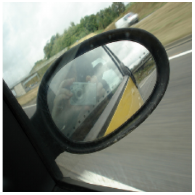}}
	{\includegraphics[]{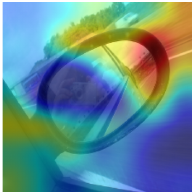}}
	{\includegraphics[]{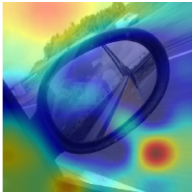}}
	{\includegraphics[]{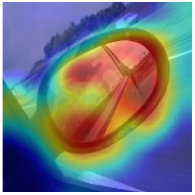}}	
	
	{\includegraphics[]{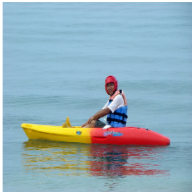}}
	{\includegraphics[]{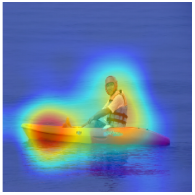}}
	{\includegraphics[]{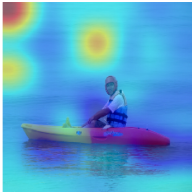}}
	{\includegraphics[]{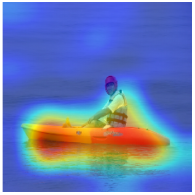}}
	{\includegraphics[]{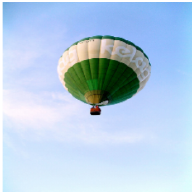}}
	{\includegraphics[]{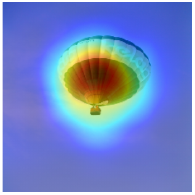}}
	{\includegraphics[]{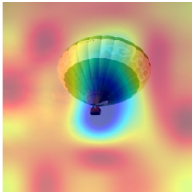}}
	{\includegraphics[]{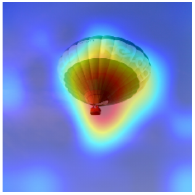}}
	

	{\includegraphics[]{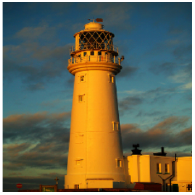}}
	{\includegraphics[]{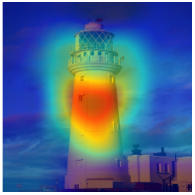}}
	{\includegraphics[]{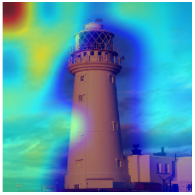}}
	{\includegraphics[]{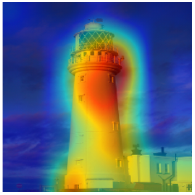}}
	{\includegraphics[]{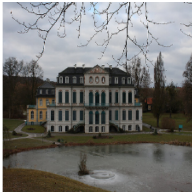}}
	{\includegraphics[]{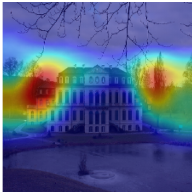}}
	{\includegraphics[]{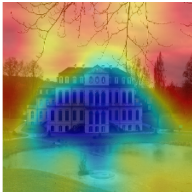}}
	{\includegraphics[]{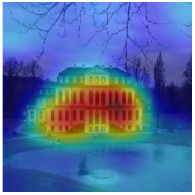}}

	{\includegraphics[]{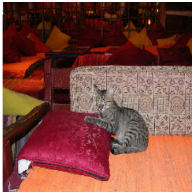}}
	{\includegraphics[]{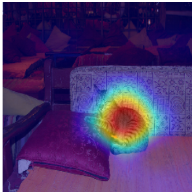}}
	{\includegraphics[]{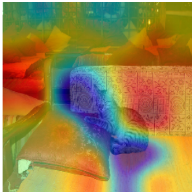}}
	{\includegraphics[]{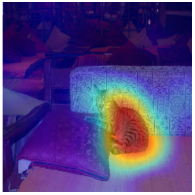}}
	{\includegraphics[]{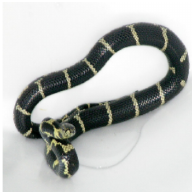}}
	{\includegraphics[]{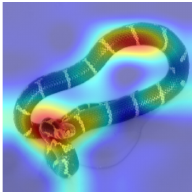}}
	{\includegraphics[]{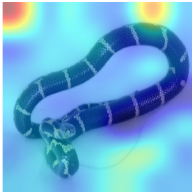}}
	{\includegraphics[]{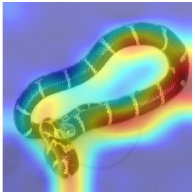}}
	\caption{Comparison of heatmaps generated by CAM with CNN, CAM with ViG, and VS-CAM. Input image: trifle, car wheel, canoe, beacon,  Egyptian cat (first column), respirator, car mirror, balloon, castle, and king snake (fifth column). Normal CAM with CNN (the second and the sixth columns). Normal CAM with ViG (the third and the seventh columns). VS-CAM (the fourth and the eighth columns).}\label{fig:comparisoncam2}
\end{figure*}

Fig.~\ref{fig: camcomparison} has shown the comparison of saliency heatmaps generated by CAM with CNN, CAM with ViG, and VS-CAM. Obviously, the highlighted regions in VS-CAM heatmaps match the object more precisely than the others. 
We further provide more results of a broader range of ten categories of images (including food, animal, architecture, transportation, daily staff, etc.), as shown in Fig.~\ref{fig:comparisoncam2}. Obviously, VS-CAM can highlight the object-relevant regions more precisely than normal Grad-CAM with CNN and ViG.

\subsection{Interpretation of ViG's Decisions}
To further understand how features are merged in graph topology before the classification layer,
we adopt three measurements, intersection angle, projection, and inner product, to evaluate the similarity among vertices in the last block of ViG.

For two vertices, $\mathbf{v}_1$, $\mathbf{v}_2\in\mathbb{R}^{1\times D}$, 
the inner product is
\begin{align}
	(\mathbf{v}_{1}, \mathbf{v}_{2})	= v_{11}v_{21} + v_{12}v_{22} + \dots +  v_{1D}v_{2D} ,
\end{align}
the intersection angle, $\angle ( \mathbf{v}_1, \mathbf{v}_2)$, is:
\begin{equation}
	\angle ( \mathbf{v}_1, \mathbf{v}_2) = \frac{(\mathbf{v}_1,\mathbf{v}_2)}{|\mathbf{v}_1||\mathbf{v}_2|},
\end{equation}
the projection is:
\begin{equation}
	\mathrm{projection}_{\mathbf{v}_2}(\mathbf{v}_1) = \frac{(\mathbf{v}_1,\mathbf{v}_2)}{|\mathbf{v}_2|},
\end{equation}
where projection is not a commutative operation.

\begin{figure}[t]
	\centering
	{\includegraphics[width=2cm]{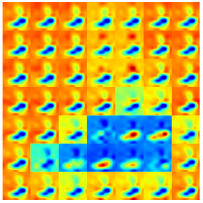}}
	{\includegraphics[width=2cm]{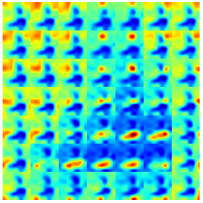}}
	{\includegraphics[width=2cm]{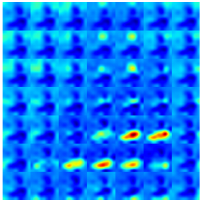}}
	{\includegraphics[width=2cm]{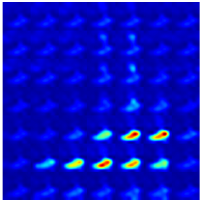}}
	
	{\includegraphics[width=2cm]{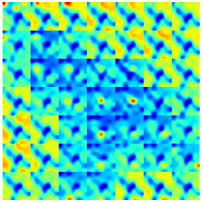}}
	{\includegraphics[width=2cm]{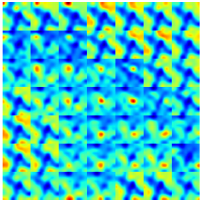}}
	{\includegraphics[width=2cm]{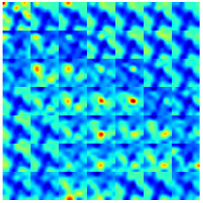}}
	{\includegraphics[width=2cm]{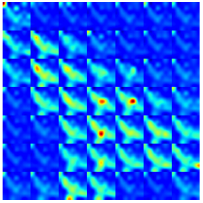}}

	{\includegraphics[width=2cm]{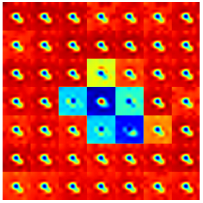}}
	{\includegraphics[width=2cm]{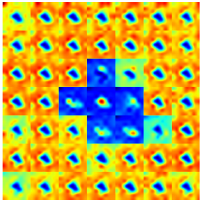}}
	{\includegraphics[width=2cm]{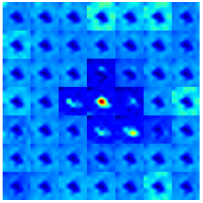}}
	{\includegraphics[width=2cm]{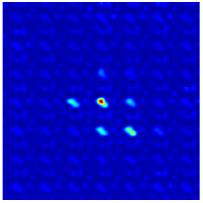}}
	
	{\includegraphics[width=2cm]{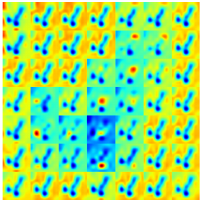}}
	{\includegraphics[width=2cm]{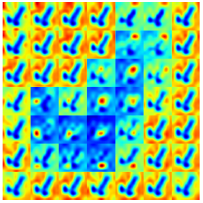}}
	{\includegraphics[width=2cm]{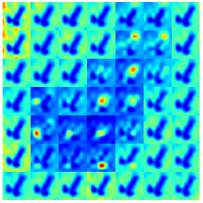}}
	{\includegraphics[width=2cm]{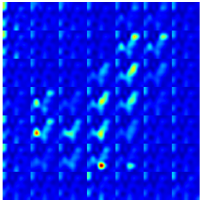}}
	
	\caption{Semantics-probe Maps corresponding to the vertices of the last block in ViG. The shape features maps of the Grapher in $12$-th block is $\mathbb{R}^{7\times7\times384}$, denoting there are $7\times7=49$ vertices and each is a vector in the length of $384$. According to (\ref{eq:similarity_matrix}), there are $49$ similarity matrices ($49$ patches in each subfigure) corresponding to $49$ vertices and each similarity matrix $S \in \mathbb{R}^{7\times7}$ ($49$ pixels in each patch). Similarity measured by the Euclidean distance between the current vertex and other vertices (first column). Similarity measured by the intersection angle. (second column)  The corresponding input images. (third column) Similarity measured by the inner product (fourth column). 
	 }\label{fig:topology12}
\end{figure}

Fig.~\ref{fig:topology12} shows the semantic-probe maps in the $12$-th block of ViG,  evaluated by Euclidean distance, intersection angle, projection, and inner product, respectively. Note there are $7\times7$ vertices and each vertex has a similarity matrix in the shape of $7\times7$, thus $49$ semantic-probe maps are in one subfigure and each semantic-probe map contains $49$ elements in the first three columns in Fig.~\ref{fig:topology12}. 
The first column shows that the current vertex has a shorter distance than those of similar nature.
The second column only uses the cosine of intersection angle, so the similarity is maximal when the vertex itself is compared. Intersection angle does not consider any intensity information of two vertices, thus the highlighted regions always appear around the current vertex itself in each sub-figure no matter whether it is object-relevant or not.
The third column further preserves the energy of the current vector, thus the pixels relevant to the object are strongly highlighted. It is because projection only retains the intensity of the current vertex itself, so the connections among object-irrelevant vertices are dramatically suppressed in comparison to object-relevant vertices. 
The fourth column uses the complete inner product and only the pixels mostly relevant to the object are highlighted. Intuitively, the highlighted region in the whole plane looks like the shape of the object roughly.
Furthermore, it can be observed two interesting phenomena in Fig.~\ref{fig:topology12}: (1) the highlighted semantic-probe map constitutes the basic profile of the object, demonstrating the object-relevant vertices are much stronger than other object-irrelevant vertices; (2) the highlighted elements in these semantic-probe maps in (1) also constitutes the object shape, indicating that an object-relevant vertex almost only connects to other object-relevant vertices rather than object-irrelevant vertices. These results illustrate that ViG can extract precise and abundant class-discriminative features and merge them into deep layers. It is the reason why semantic-probe maps work effectively to detect semantics.

\subsection{Topology of Vertices in ViG}
\begin{figure}[t]
	\centering
	{\includegraphics[width=1.3cm]{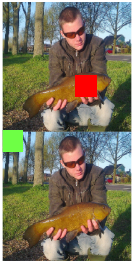}}
	{\includegraphics[width=1.3cm]{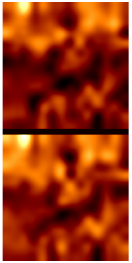}}
	{\includegraphics[width=1.3cm]{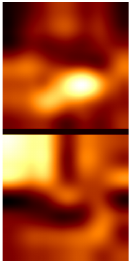}}
	{\includegraphics[width=1.3cm]{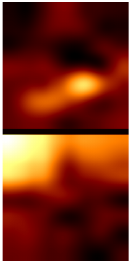}}
	{\includegraphics[width=1.3cm]{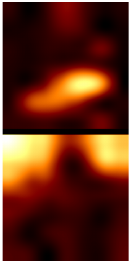}}
	{\includegraphics[width=1.3cm]{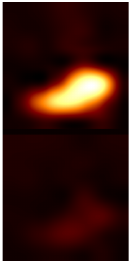}}
	
	{\includegraphics[width=1.3cm]{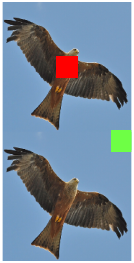}}
	{\includegraphics[width=1.3cm]{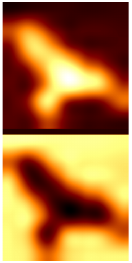}}
	{\includegraphics[width=1.3cm]{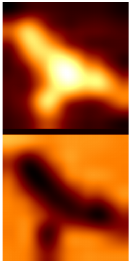}}
	{\includegraphics[width=1.3cm]{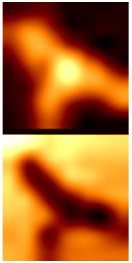}}
	{\includegraphics[width=1.3cm]{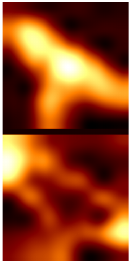}}
	{\includegraphics[width=1.3cm]{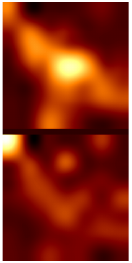}}
	{\includegraphics[width=1.3cm]{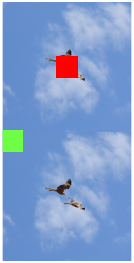}}
	{\includegraphics[width=1.3cm]{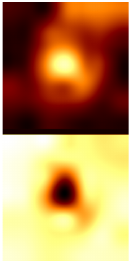}}
	{\includegraphics[width=1.3cm]{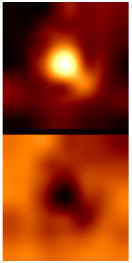}}
	{\includegraphics[width=1.3cm]{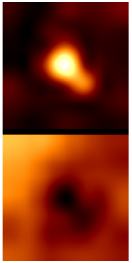}}
	{\includegraphics[width=1.3cm]{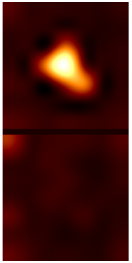}}
	{\includegraphics[width=1.3cm]{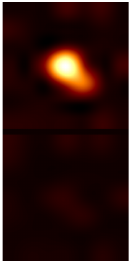}}

	{\includegraphics[width=1.3cm]{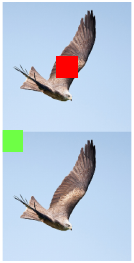}}
	{\includegraphics[width=1.3cm]{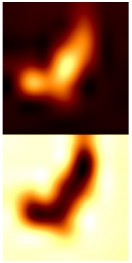}}
	{\includegraphics[width=1.3cm]{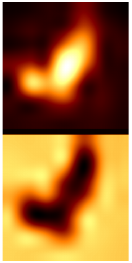}}
	{\includegraphics[width=1.3cm]{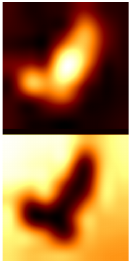}}
	{\includegraphics[width=1.3cm]{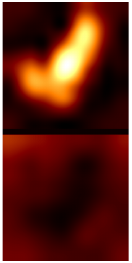}}
	{\includegraphics[width=1.3cm]{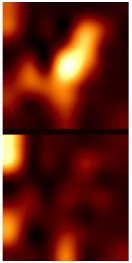}}
	\caption{The topology of graph connections to vertices in different blocks of ViG. We select two vertices, $\mathbf{v}_1$ and $v_2$ ($v_1$ is highly relevant to the object while $v_2$ is not.) from the output $\in\mathbb{R}^{7\times7\times 384}$  of $12$-th block in ViG. Then we can obtain the topology connections between two selected vertices and the rest. The highlighted regions mean where the current vertex is strongly connected to. Following the above operation, we further visualize the topology of such vertices in different blocks ($l$ = $2$, $4$, $5$, $10$). Note that the output of each block is downsampled to $7\times7\times D^{l}$ by average pooling operator to ensure the same coordinates of $v_1$ and $v_2$ in $12$-th block.  The region corresponding to the selected two vertices is also marked in the input images (red square denotes $v_1$ and green square denotes $v_2$). }\label{fig:topology}
\end{figure}
To fully understand the mechanism of ViG,
we also visualize the semantic-probe maps corresponding to all vertices in different blocks in Fig.~\ref{fig:topology}. Two
center vertices, $\mathbf{v}_1$ and $\mathbf{v}_2$ ($\mathbf{v}_1$ is an object-relevant vertex and $\mathbf{v}_2$ is an object-irrelevant vertex), are visualized as drawing all the vertices will be messy, especially in shallow blocks. Note the output shape of $12$-th block is much smaller than the input shape, thus $\mathbf{v}_{1}$ and $\mathbf{v}_2$ are mapped to a scaled square in the input image ($\mathbf{v}_1$ and $\mathbf{v}_2$ are marked with red and green in Fig.~\ref{fig:topology} (fourth column), respectively). 
It is clearly observed that $\mathbf{v}_1$ almost only connects to other object-relevant vertices while $\mathbf{v}_2$ attempts to avoid such connections. Furthermore, in shallow blocks, some object-irrelevant vertices close to the object are still weakly connected to $\mathbf{v}_1$ (especially the Tench with a complex background in the second row in Fig.~\ref{fig:topology}), while it can not be observed in deep blocks. It demonstrates that ViG tends to connect vertices based on low-level and local features, such as color and texture in shallow blocks, especially the input image with a complex background, like Fig.~\ref{fig:topology}(the first row). In contrast, almost only those features are more semantic and category-discriminative are fused in deep blocks.

\subsection{Ablation Study}

In this section, we study how some factors affect VS-CAM, like the selection of similarity measurement (intersection angle, projection, inner product) and the number of semantic-probe maps.

\begin{figure}[t]
	\centering
	{\includegraphics{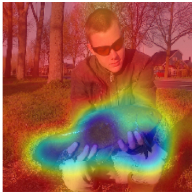}}
	{\includegraphics{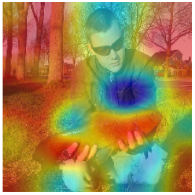}}
	{\includegraphics{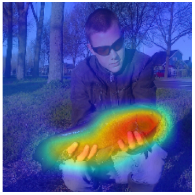}}
	{\includegraphics{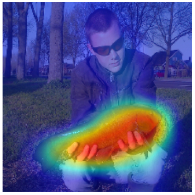}}
	
	{\includegraphics{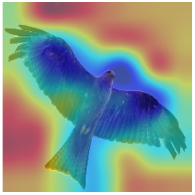}}
	{\includegraphics{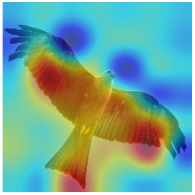}}
	{\includegraphics{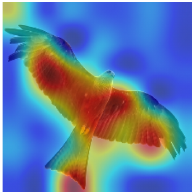}}
	{\includegraphics{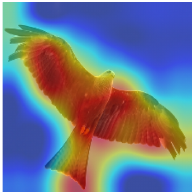}}
	
	{\includegraphics{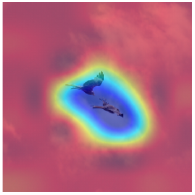}}
	{\includegraphics{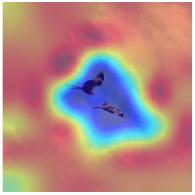}}
	{\includegraphics{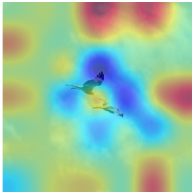}}
	{\includegraphics{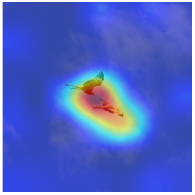}}
	
	{\includegraphics{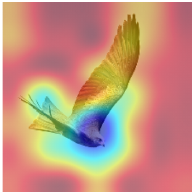}}
	{\includegraphics{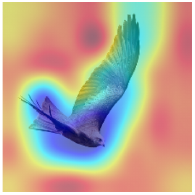}}
	{\includegraphics{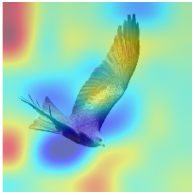}}
	{\includegraphics{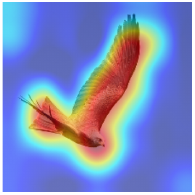}}
	\caption{Effect of different similarity measurements on VS-CAM with above four images. From left to right subfigures are obtained by using Euclidean distance (the first column) intersection angle  (the second column), projection  (the third column), and inner product  (the fourth column), respectively.  }\label{fig:ablation_similarity}
\end{figure}

\noindent\textbf{Effect of Similarity Measurement} We study how different similarity measurements affect the heatmaps of VS-CAM as discussed in Sec.~\ref{sec:3}. Here we present the heatmaps by using intersection angle, projection, and inner product in Fig.~\ref{fig:ablation_similarity}, respectively. The results show that intersection angle and projection may cause odd and intricate highlighted regions as well as some semantic reversal for some objects. It demonstrates that introducing the intensity of both vertices is necessary for similarity measurement and it is the reason why the inner product is chosen in VS-CAM.


\begin{figure}[t]
	\centering
	{\includegraphics{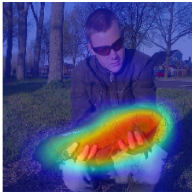}}
	{\includegraphics{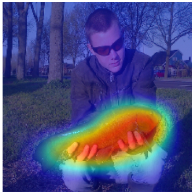}}
	{\includegraphics{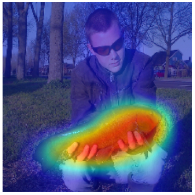}}
	{\includegraphics{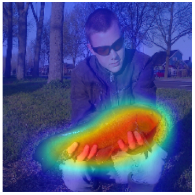}}	
	{\includegraphics{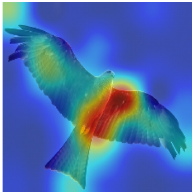}}
	{\includegraphics{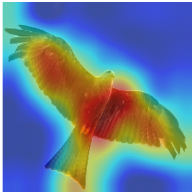}}
	{\includegraphics{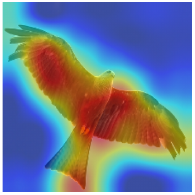}}
	{\includegraphics{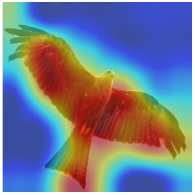}}	
	{\includegraphics{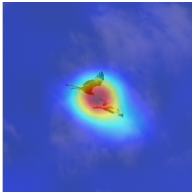}}
	{\includegraphics{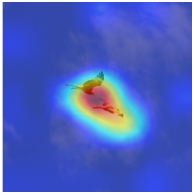}}
	{\includegraphics{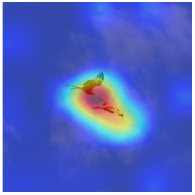}}
	{\includegraphics{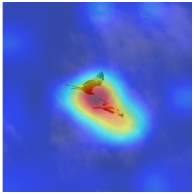}}	
	{\includegraphics{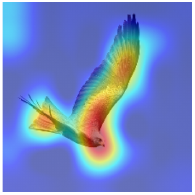}}
	{\includegraphics{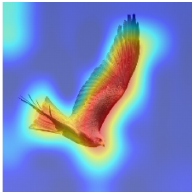}}
	{\includegraphics{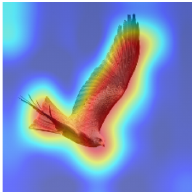}}
	{\includegraphics{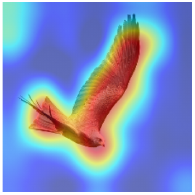}}
	\caption{Effect of the number of semantic-probe maps. From left to right subfigures are obtained with the number of semantic-probe maps: $1$ (the first column), $7$ (the second column), $14$  (the third column), and $49$ (the fourth column), respectively.  }	
	\label{fig:ablation_numknn} 
\end{figure}

\noindent\textbf{Number of Semantic-probe Maps} In previous experiments, all the semantic-probe maps are used, however, it is probable that we can avoid some unimportant maps to be more efficient. Thus we only select the first several maximal semantic-probe maps to see the change of heatmaps, as shown in Fig.~\ref{fig:ablation_numknn}. We can see that VS-CAM performs well when the number of semantic-probe maps is over $14$, while it does not work well with a small number of semantic-probe maps.

\subsection{Quantitative Analysis}

\begin{figure}[t]
	\centering
	{\includegraphics{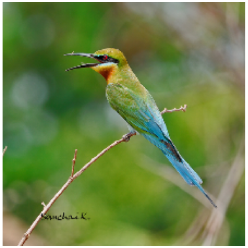}}
	{\includegraphics{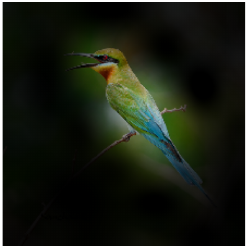}}
	{\includegraphics{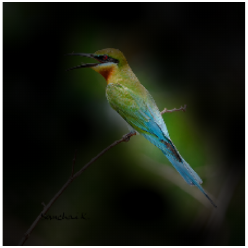}}
	{\includegraphics{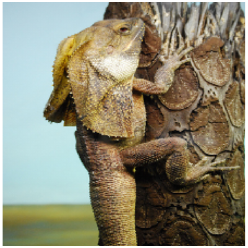}}
	{\includegraphics{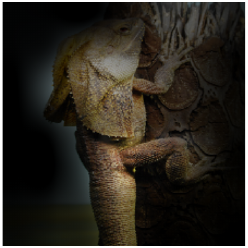}}
	{\includegraphics{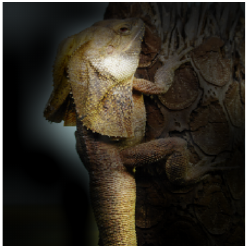}}
	\caption{The explanation maps of Grad-CAM and VS-CAM. The first column is the original images. The second column is the explanation maps generated by Grad-CAM. The third column is the explanation maps generated by VS-CAM.}\label{fig: explanationmap} 
\end{figure}

\begin{figure}[t]
	\centering
	{\includegraphics[width=4cm]{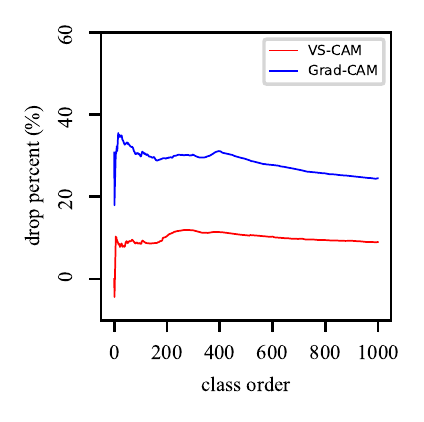}}
	{\includegraphics[width=4cm]{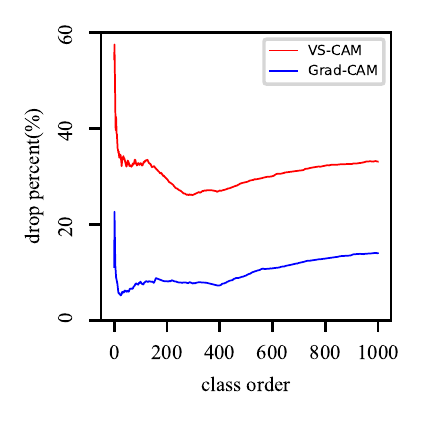}}
	{\includegraphics[width=4cm]{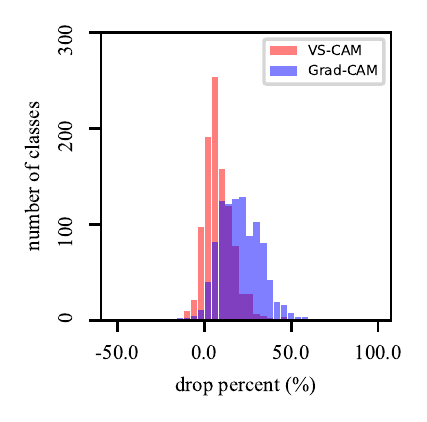}}
	{\includegraphics[width=4cm]{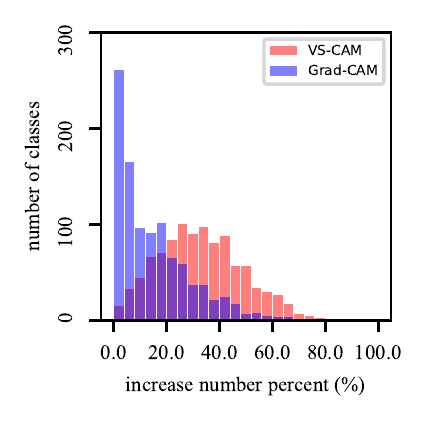}}
	\caption{The confidence drop and increase number of Grad-CAM and VS-CAM in the entire dataset. The curve of confidence drop in the entire validation set (top-left).  The curve of increase number in the entire validation set (top-right). The histogram of confidence drop in each class (bottom-left). The histogram of increase number in each class (bottom-right).}\label{fig: curve} 
\end{figure}


To further analyze the interpreting performance of VS-CAM quantitatively, we utilize two commonly-used evaluation metrics, i.e., confidence drop and confidence increase \cite{GradCAMplus}.
First of all, we need to consider what kind of heatmap can be deemed to provide a good interpretation of the neural network. A natural and intuitive idea is to measure how much the confidence of the correct class will drop when the original image is partly occluded according to the heatmaps. Specifically, 
for each image, a corresponding explanation map $\mathbf{L}_c$ is generated
by element-wise multiplication of the heatmaps and the current image as:
\begin{equation}
	\mathbf{L}_c = \mathbf{M}_c \odot \mathbf{I}.
\end{equation}
In Fig.~\ref{fig: explanationmap}, we select two images to exhibit their explanation maps (Grad-CAM with ViG will not be discussed here due to semantic-chaos). 
Then we can define two evaluation metrics.

\noindent\textbf{Confidence drop}: This metric compares the average drop of
the model’s confidence for a particular class in an image
after occlusion as:
\begin{equation}
	\mathrm{confidence\_drop} =	\frac{\Im(\mathbf{I}) - \Im(\mathbf{L}_c)}{\Im(\mathbf{I})},
\end{equation}
For instance, suppose that the model predicts
an object "Tench" in an image $\mathbf{I}$ with confidence $0.8$. When
shown the explanation map, $\mathbf{L}_c$, of this image, the model’s
confidence in the class "Tench" falls to $0.2$. Then the $\mathrm{confidence\_drop}$ would be $75\%$. It means that only a small part of the object is in the discriminative highlighted region in the heatmaps. This value is averaged over the entire dataset. 

\noindent\textbf{Increase number}: This metric
measures the number of times the
model’s prediction score for $c$ increased in the entire dataset. Specifically, sometimes it is
possible that the entire object is included  and other interference parts are occluded (e.g., the object-irrelevant parts and background) in the most discriminative part highlighted by the
explanation maps. In this scenario, there is an increase in
the model’s confidence for that particular class (i.e., $\mathrm{confidence\_drop}$  is a negative value).  This value is expressed as a percentage.

Table~\ref{tab:metric} shows the two evaluation metrics of the entire validation set in ILSVRC dataset ($\downarrow$ means lower value is better and $\uparrow$ means higher value is better).  These results clearly indicate a superior performance of the proposed VS-CAM to Grad-CAM. Furthermore, we present the curves of two metrics in the entire validation set and the histograms of each class in Fig.~\ref{fig: curve}. The curves show that VS-CAM always outperforms the Grad-CAM in both evaluation metrics. The histogram of confidence drop shows that most classes of images occluded by VS-CAM obtain lower confidence drop than Grad-CAM.  The histogram of increase number shows that the prediction score of many classes  is improved after occluded by VS-CAM while a much fewer happened when Grad-CAM is adopted. Above metrics values are computed in Pytorch 1.8.0+cudnn11.1, NVIDA RTX-3070.

\begin{table}[tbp]
	\centering
	\caption{Evaluation  Metrics. \label{tab:metric}}
	\begin{tabular}{ccccc}
		\toprule  
		Method&Confidence drop $\downarrow$&Increase number $\uparrow$&\\ 
		\midrule  
		
		Grad-CAM&$24.49$ 	&13.97&  \\	
		VS-CAM&$\mathbf{9.01}$ 	& $\mathbf{33.05}$ &  \\	
		\bottomrule  
	\end{tabular}
\end{table}	

\section{Conclusion}\label{sec:5}
In this paper, we present a novel visualization method termed VS-CAM for graph neural network, inspired by the connections among vertices in graph topology. VS-CAM is the first attempt to interpret graph neural convolutional network in image classification tasks. VS-CAM shows that the vertices manage to build connections to their homogeneity and avoid connecting to irrelevant vertices. This phenomenon is more significant in the deep layer than shallow layers (i.e., the deep layers of GCN aggregate more senior semantic features). Experimental results demonstrate the validity and superiority of VS-CAM to other comparative CAM methods.

\section*{Data Availability Statements}\label{res:sec6}
ILSVRC dataset can be downloaded from
\href{http://image-net.org/}{http://image-net.org/} and
ViG model can be downloaded from \href{https://github.com/huawei-noah/Efficient-AI-Backbones}{ViG}.

\section*{Acknowledgements}\label{res:sec7}
This work is funded by Science and technology project of Xianyang city (2021ZDZX-GY-0001), the National Natural Science Foundation of China (No. 61871301), the National Natural Science Foundation of China (No. 62071349).
The authors are thankful to Prof. Milo\v s Dakovi\'c for the help in the preparation of this manuscript.

\bibliographystyle{cas-model2-names}

\bibliography{References}

\newpage
\bio{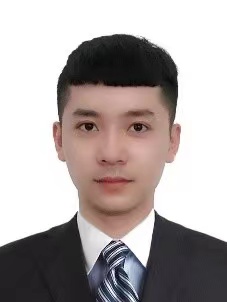}
\textbf{Zhenpeng Feng} 
 was born in Xi'an, Shaanxi, China in 1996. He received the B.E. degree in School of Electronic Engineering, Xidian University in 2019. He is currently a Ph.D student in explainable artificial intelligence with School of Electronic Engineering, Xidian University. He is also a visiting student in University of Montenegro, working with Prof. Ljubi\v{s}a Stankovi\'c's research team.  His research interests include interpreting deep networks and signal processing.
\endbio

\bio{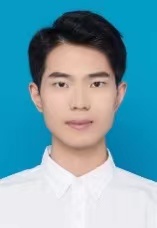}
\textbf{Xiyang Cui} 
was born in Handan, Heibei, China in 1997. He received the B.E. degree and M.E. degree in Electronic Information Engineering and Electrical Circuit System with School of Electronic Engineering, Xidian University in 2019 and 2021, respectively. He is currently an investigator of an electronic company and collaborate with Zhenpeng Feng and Prof. Ljubi\v sa Stankovi\'c in scientific research.  His research interests include electrical circuit design and image processing.
\endbio

\bio{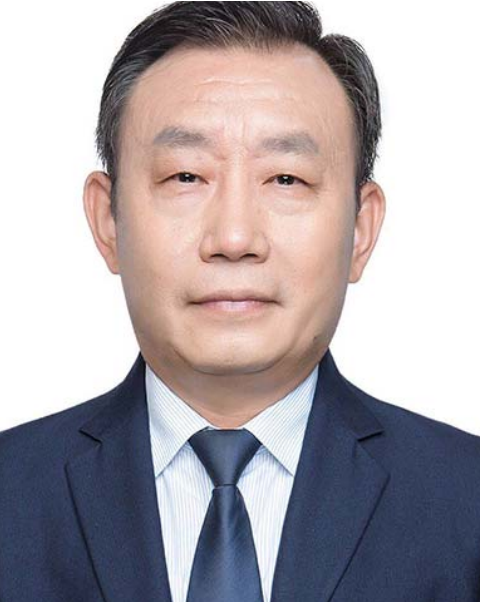}
\textbf{Hongbing Ji} 
received the B.S. degree in radar engineering, the M.S. degree in
circuit, signals and systems, and the Ph.D. degree in
signal and information processing from
Xidian University, Xi’an, China, in 1983,
1989, and 1999, respectively. He is currently a full professor in Xidian University. His research interests
include pattern recognition, radar signal processing,
and multi-sensor information fusion.
\endbio

\bio{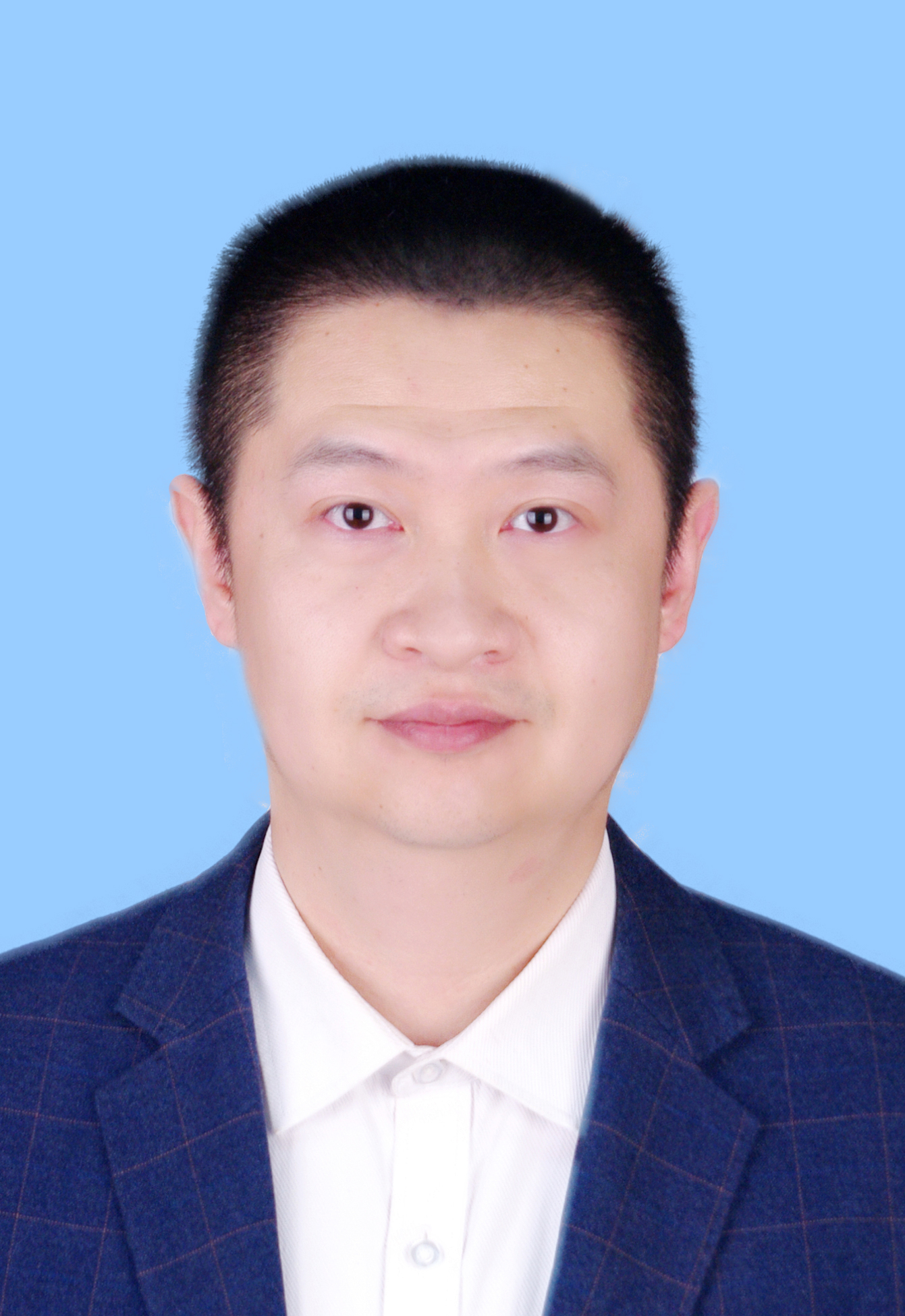}
\textbf{Mingzhe Zhu} 
was born in China in 1982.
He received the B.S. degree in signal and information processing, the Ph.D. degree in
pattern recognition and intelligent system from Xidian University in 2004 and 2010, respectively. He is currently an associate professor with the School of Electronic Engineering, Xidian University. His
research interests include non-stationary
signal processing, time-frequency analysis
and target recognition.
\endbio

\bio{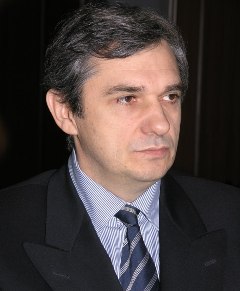}
\textbf{Ljubi\v{s}a Stankovi\'{c}}  was born in Montenegro, 1960. He was at the Ruhr University Bochum, 1997-1999, supported by the AvH Foundation. Stankovic was the Rector of the University of Montenegro 2003-2008, the Ambassador of Montenegro to the UK, 2011-2015 and a visiting academic to the Imperial College London, 2012-2013.  He published almost journal 200 papers. He is a member of the National Academy of Science and Arts (CANU) and of the Academia Europaea. Stankovic won the Best paper award from the EURASIP in 2017 and the IEEE SPM Best Column Award for 2020. Stankovic is a professor at the University of Montenegro and a Fellow of the IEEE.
\endbio

\end{document}